\documentclass[]{fairmeta}

\usepackage{listings}

\lstdefinestyle{filesystem}{
    basicstyle=\ttfamily\small,
    backgroundcolor=\color{gray!6},
    frame=single,
    rulecolor=\color{gray!35},
    framesep=8pt,
    xleftmargin=4pt,
    xrightmargin=4pt,
    columns=fullflexible,
    keepspaces=true,
    showstringspaces=false,
    breaklines=true,
    aboveskip=8pt,
    belowskip=8pt
}

\title{Auto-RecSys: Harnessing Autonomous Research Agents for Industry-Scale Recommender Systems}

\author[1]{Ming Li}
\author[1]{Dai Li}
\author[1,2]{Xuying Ning}
\author[1]{Bo Sun}
\author[1]{Rui Li}
\author[1]{Yi Zhang}
\author[1]{Silvia Gong}
\author[1]{Xuan Cao}
\author[1]{Rui Li}
\author[1]{Cornelia Carapcea}
\author[1]{Qunshu Zhang}
\author[1]{Zhigang Wang}
\author[1]{Yinglong Xia}
\author[1]{Xue Feng}
\author[1]{Andy Wang}

\affiliation[1]{Meta}
\affiliation[2]{University of Illinois Urbana-Champaign}

\abstract{

Auto-research agents have shown the potential to automate hypothesis generation, experiment execution, and iterative refinement. However, scaling this paradigm to industry-scale recommendation models introduces two challenges: (1) \emph{long feedback loops}, where model training can take days, making serial iteration prohibitively slow and requiring parallel exploration across multiple research directions; and (2) \emph{system complexity}, where large configurations, fragile infrastructure dependencies, and multi-day GPU jobs require robust and recoverable execution. We present \textbf{Auto-RecSys}, an autonomous research system for long-horizon experimentation on industry-scale recommendation models. Auto-RecSys addresses these challenges through three harness designs: (1) distributed asynchronous execution for running multiple experiments in parallel across servers, (2) centralized cross-server memory for persistent and recoverable execution across sessions and failures, and (3) cognitive-procedural separation, where natural-language skill files guide LLM reasoning while deterministic scripts enforce operational correctness. Auto-RecSys further employs a \emph{dual-loop self-evolving architecture}: an Execution Evolution Loop in which model-specific playbooks accumulate operational knowledge by recording failed attempts and crystallizing successful pipelines, and an Idea Evolution Loop in which experimental outcomes inform subsequent ideation. Evaluated on recommendation models, Auto-RecSys significantly reduces the human time required per experiment cycle and improves execution reliability as its playbooks mature.

}

\date{\today}
\correspondence{Andy Wang at \email{xiaolongw@meta.com}; Ming Li at \email{liming@meta.com}} 

\begin{document}

\maketitle

\section{Introduction}
\label{sec:intro}

The vision of autonomous AI research, in which agents generate hypotheses, execute experiments, analyze results, and iteratively refine their ideas, is rapidly becoming a reality~\citep{chen2025ai4research}. Recent systems have demonstrated that LLM-powered agents can coordinate increasingly complete research workflows, including hypothesis generation, code implementation, empirical evaluation, and theory refinement~\citep{aiscientist,evoscientist2026}. AutoResearch supports high-throughput experimentation on small-scale problems~\citep{autoresearch}, while FARS demonstrates that autonomous agents can produce research artifacts at scale on bounded tasks~\citep{fars}. These advances establish the potential of automated research when experiments are self-contained and feedback is available within minutes or hours.

Applying this paradigm to industry-scale recommendation models introduces a different operating regime. A single model change may require days of training and monitoring, while implementation depends on large configuration stacks, distributed infrastructure, and long-running GPU jobs. A typical research cycle spans ideation, implementation, validation, training, failure recovery, and result analysis, often taking three to seven days. Researchers therefore spend substantial effort managing execution rather than developing and evaluating new ideas. In this setting, autonomous research is not only a question of whether an agent can propose promising model changes. It also requires a system that can explore multiple directions concurrently, execute experiments reliably, recover from interruptions, and retain operational knowledge across research cycles. As summarized in \Cref{table:comparison}, this setting introduces two challenges.

\paragraph{Challenge 1: Long feedback loops require parallel exploration.}
Existing autonomous research systems can iterate serially because experiments complete quickly: an agent proposes an idea, evaluates it, analyzes the result, and begins the next iteration. For industry-scale recommendation models, a single training run can consume days of GPU time. Serial experimentation at this timescale severely limits research throughput. Maintaining meaningful research velocity therefore requires multiple ideas to be implemented, trained, and monitored concurrently across distributed resources, with results incorporated asynchronously as they become available.

\paragraph{Challenge 2: System complexity requires robust and persistent execution.}
Small-scale research environments are often self-contained and relatively easy to reproduce. Industry-scale recommendation models instead involve thousands of lines of configuration, dependencies across code and infrastructure, specialized hardware requirements, and jobs that may fail because of preemption, checkpoint corruption, stale data, or compatibility issues. Agent sessions may terminate during implementation, servers may restart during training, and experiments may move across development environments. Because every run is computationally expensive, execution must remain recoverable across sessions, servers, and failures. The system must also preserve successful procedures, failed attempts, and their underlying causes so that operational knowledge is accumulated rather than repeatedly rediscovered.

\begin{table*}[t]
    \centering
    \setlength{\tabcolsep}{3pt}
    \begin{NiceTabular}{lll}
        \toprule
        \textbf{Dimension} &
        \textbf{Small-Scale Auto-Research} &
        \textbf{Industry-Scale Auto-RecSys} \\
        \midrule
        Feedback loop
        & Minutes to hours
        & Hours to days \\
        Iteration strategy
        & Rapid serial iteration
        & Distributed parallel exploration \\
        Computational cost
        & Relatively low
        & High \\
        Failure recovery
        & Local reruns
        & Persistent recovery across sessions \\
        Operational complexity
        & Self-contained code
        & Large configurations and infrastructure dependencies \\
        Agent lifetime
        & Single continuous session
        & Multiple sessions, servers, and asynchronous stages \\
        \bottomrule
    \end{NiceTabular}
    \caption{
    Comparison between small-scale autonomous research and automated research
    on industry-scale recommendation models. Long feedback loops, computationally expensive
    experiments, and operational complexity change the required
    system design.
    }
    \label{table:comparison}
\end{table*}

We present \textbf{Auto-RecSys}, an autonomous research system for long-horizon experimentation on industry-scale recommendation models. Auto-RecSys supports the full experiment lifecycle, from idea formulation and implementation to validation, training, monitoring, debugging, and result analysis. To make this lifecycle reliable under long training windows and distributed execution, Auto-RecSys is built around an agent harness that manages how research agents schedule experiments, preserve state, interact with infrastructure, recover from failures, and reuse accumulated execution knowledge. The harness is organized around three designs:

\begin{itemize}
    \item \textbf{Distributed asynchronous execution.}
    Multiple experiment ideas proceed concurrently across distributed servers and may occupy different stages of the research lifecycle. A persistent state machine tracks each experiment independently, allowing agents to coordinate work across multi-day training windows without relying on a single continuous session.

    \item \textbf{Centralized cross-server memory.}
    Experiment states, model-specific playbooks, execution histories, and session trajectories are stored in a shared memory layer accessible from different development servers. When an agent session terminates or a server restarts, another session can reconstruct the experiment context and resume execution from the persisted state.

    \item \textbf{Cognitive-procedural separation.}
    Natural-language skill files guide agent reasoning, planning, and diagnosis, while deterministic scripts perform state transitions, API interactions, validation checks, and file operations. This separation preserves the flexibility of LLM reasoning while enforcing correctness for operationally sensitive actions.
\end{itemize}

\begin{figure*}[t]
    \centering
    \includegraphics[width=0.75\textwidth]{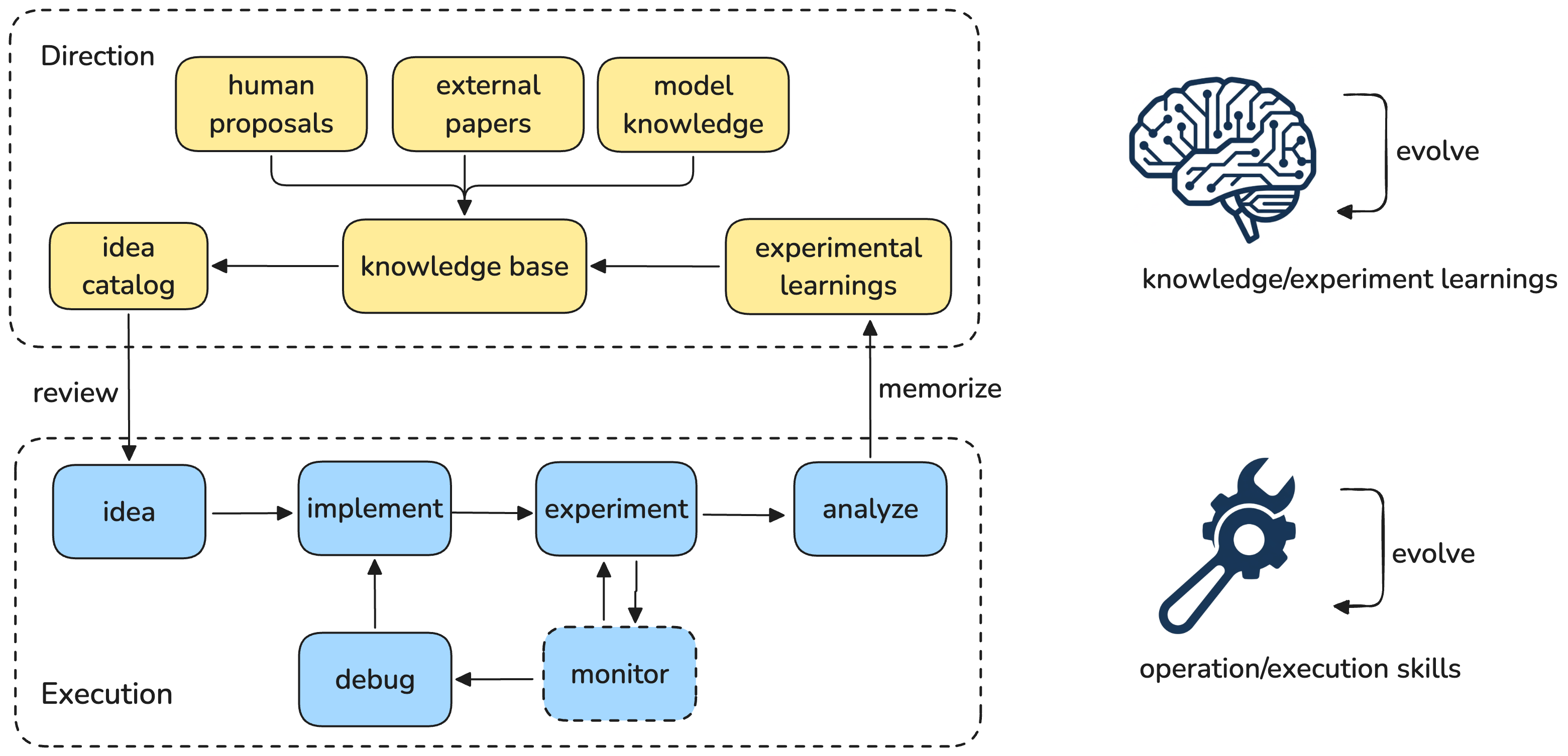}
    \caption{
    Overview of the dual-loop self-evolving architecture in Auto-RecSys.
    The Idea Evolution Loop feeds experiment outcomes into future proposal
    generation, while the Execution Evolution Loop distills execution
    trajectories into model-specific playbooks containing validated workflows
    and known dead ends.
    }
    \label{fig:system-overview}
\end{figure*}

Auto-RecSys further improves through a \textbf{dual-loop self-evolving architecture}, illustrated in \Cref{fig:system-overview}. The \emph{Execution Evolution Loop} distills experiment trajectories into model-specific playbooks that record relevant files, validated commands, hardware requirements, recurring failures, and successful pipeline configurations. Failed attempts are retained as dead ends, while successful procedures are crystallized into reusable workflows, allowing execution to become increasingly reliable as the playbooks mature. In parallel, the \emph{Idea Evolution Loop} records experimental outcomes and scientific conclusions so that future proposals can incorporate prior evidence, avoid redundant directions, and adapt to changes in model baselines. Together, the two loops allow Auto-RecSys to improve both how experiments are executed and which ideas are explored.

We evaluate Auto-RecSys on recommendation models, where it significantly reduces the human time required per experiment cycle and measurably improves execution reliability as its model-specific playbooks mature. These results demonstrate that autonomous research can be extended beyond short, self-contained experiments to long-horizon, infrastructure-intensive model development through a harness designed for parallel execution, persistent recovery, and continual knowledge accumulation.

\section{Auto-RecSys Overview}
\label{sec:overview}

Auto-RecSys organizes the full experiment lifecycle around a layered design. An orchestration layer drives a finite state machine and routes work to specialist agents. A specialist agent layer handles each stage of the research loop, from ideation through analysis, with one agent per state of the experiment lifecycle. A persistence layer stores experiment state, model playbooks, and experiment history in a shared memory store that every development server can reach. This section describes the principles behind this organization and the harness that implements it.

\subsection{Design Principles}

Five principles guide the design of Auto-RecSys, each motivated by the demands of operating at industry scale. First, the system composes rather than reinvents. It builds on existing infrastructure for distributed training, job submission, version control, and metric monitoring, and each specialist agent is a thin orchestration layer over these proven tools. Composability matters in a setting where infrastructure changes frequently and reliability cannot be compromised. Second, the system keeps a human in the loop at decision points. In interactive mode it pauses at four checkpoints, namely idea selection, code review, training submission, and results review, whereas in autonomous mode these checkpoints are skipped; autonomous operation becomes progressively safer as playbook confidence grows. Third, the system is asynchronous by design. It never assumes that the same agent session, or even the same server, will remain available when a multi-day training job completes. Fourth, the system is self-evolving: both execution procedures and research directions improve over time through structured feedback rather than manual rule authoring, so operational knowledge becomes an appreciating asset rather than a static configuration. Finally, the system treats failure as expected and recoverable, relying on atomic writes to prevent state corruption, append-only logs to preserve history, trajectory logs to enable context recovery, and a catalog of known dead ends to avoid repeating computationally expensive mistakes.

\subsection{Harness Design}
\label{sec:harness}

Auto-RecSys is built on an LLM agent harness (i.e., the code and context that wraps a language model and determines what information it sees at each step)~\citep{lee2026metaharness}. We organize this harness around three architectural patterns, each addressing a distinct challenge of deploying LLM agents for long-running, stateful tasks.

\paragraph{Cognitive-Procedural Separation.} Following the reasoning-acting paradigm of ReAct~\citep{yao2023react} and the agent-computer interface design of SWE-agent~\citep{yang2024sweagent}, Auto-RecSys separates the agent into two complementary layers that map to the code-vs-natural-language harness duality identified by Ning et al.~\citep{ning2026codeharness}. The \emph{natural-language harness layer} consists of skill files that describe \emph{what} to do at each stage, \emph{why} certain decisions matter, and \emph{when} to escalate to the human. The LLM reads these skills and exercises its planning and reasoning capabilities to decide the next action. The \emph{code harness layer} consists of deterministic code scripts that execute precise state transitions, file writes, and API calls, etc. The LLM invokes these scripts as tool calls, and the scripts enforce preconditions, validate inputs, and guarantee atomic state updates. The separation is important because LLM reasoning is flexible but imprecise, while state management requires exactness: a single wrong field in a JSON state file can corrupt an entire experiment lifecycle and make failure recovery far harder.

\paragraph{Hierarchical Knowledge Architecture.} Inspired by the tiered memory of MemGPT~\citep{packer2023memgpt} and the skill library of Voyager~\citep{wang2023voyager}, Auto-RecSys organizes knowledge in three layers of increasing specificity. At the most general level, a model-agnostic orchestrator skill defines the core experiment loop once and shares it across all model types, capturing the general knowledge of how to run an ML experiment, including state transitions, idea selection, baseline management, and analysis methodology. Beneath it, each model type maintains its own playbook, a compact natural-language skill artifact~\citep{yang2026skillopt} that records the domain-specific knowledge varying between architectures, such as key files, configuration conventions, hardware requirements, dead ends, and proven strategies. At the most specific level, each experiment iteration keeps its own state file with the ephemeral details of the current task, including commit hashes, job identifiers, validation results, and analysis verdicts. This hierarchy enables rapid onboarding of new models: the orchestrator skill transfers immediately, and only the per-model playbook must be bootstrapped through a few interactive iterations.

\paragraph{Natural-Language Procedural Memory.} Where Reflexion~\citep{shinn2023reflexion} uses verbal self-reflections as episodic memory for short-horizon tasks, Auto-RecSys extends this principle to long-horizon, cross-session procedural memory for industry-scale operations. Operational knowledge, including dead ends, pipeline recipes, and submission configs, is stored as natural-language instructions in markdown documents that the LLM reads at session start. This aligns with SkillOpt's~\citep{yang2026skillopt} formulation of skills as ``portable natural-language artifacts that package procedures, domain heuristics, tool policies, output constraints, and failure modes, letting a frozen agent adapt through external text.'' Auto-RecSys's dead-end catalog functions as a \emph{rejected-edit buffer}, i.e., a record of approaches that failed and should not be repeated, while the pipeline recipes and submission configs represent \emph{accepted edits} that have been validated through successful iterations. This design choice, validated by examining 31 agent session transcripts (Section~\ref{sec:results}), reflects a finding independently supported by both Meta-Harness~\citep{lee2026metaharness} and SkillOpt: \emph{LLM agents consume procedural knowledge effectively when it is written in the same language they reason in}. Meta-Harness further shows that preserving full execution traces (rather than compressed scalar scores) contributes to effective harness evolution, a principle that Auto-RecSys's detailed dead-end descriptions instantiate.

\section{The Experiment State Machine}
\label{sec:state-machine}

Auto-RecSys execution includes a finite state machine that tracks each experiment idea through its lifecycle. Each idea progresses independently through the states \textsc{ideating}, \textsc{implementing}, \textsc{validating}, \textsc{training}, and \textsc{analyzing}, with a branch into a \textsc{debugging} state on training failure and a loop back from \textsc{analyzing} to \textsc{ideating} for the next iteration. The state machine enforces preconditions on each transition and triggers side effects, such as recording results to the experiment history when an idea is finalized and updating dead ends and pipeline recipes in the playbook.

\subsection{Per-Idea State Isolation}

One design choice, driven by the need to run ideas in parallel, is that each idea keeps its own state file rather than sharing a single mutable state per model. Because state is isolated in this way, multiple ideas can be in flight on the same model at once, each at a different stage and potentially on a different server. Isolation also contains failures: an idea whose training job fails after hours of GPU time affects only its own state file and never blocks or corrupts the ideas running alongside it.

\subsection{Operating Modes}

Auto-RecSys supports two operating modes that can be toggled at any point. Interactive mode pauses at each human checkpoint for approval and is used for new models, high-risk changes, or whenever the researcher wants tight control. Autonomous mode runs the full loop end to end, chaining ideas from the backlog without pausing. The move from interactive to autonomous operation is itself part of the Execution Evolution Loop: as the playbook matures and accumulates dead-end knowledge, autonomous mode becomes increasingly reliable.

\section{Execution Evolution Loop}
\label{sec:execution-evolution}

The Execution Evolution Loop addresses Challenge 2 (system complexity and robustness) by building an evolving institutional memory for each model. The abstraction is the playbook, a per-model pair of files consisting of a human-readable procedural recipe and a machine-readable iteration metadata file.

\subsection{Why Playbooks Are Necessary at Industry Scale}

In small-scale autonomous research, operational knowledge is trivial: run a script, check the loss, and repeat. At industry scale, operational knowledge is itself the bottleneck. Running a single experiment on a recommendation model requires knowing which of the many files in a model directory hold the configuration dataclass, the encoder, and the task heads; the naming conventions and insertion points for configuration flags; the exact validation command and what its successful output looks like; which GPU generation is stable for the model; which package layer versions are compatible with the current codebase revision; which resource entitlements and scheduling tags a training job requires; and which warm-start checkpoint is current and valid. A human researcher accumulates this knowledge over weeks. Auto-RecSys captures it in the playbook after one or two interactive iterations and makes it immediately available to all subsequent iterations, including autonomous ones.

\subsection{Playbook Structure}

Each model's playbook records six categories of operational knowledge. It captures the key configuration, model, and trainer files together with their important classes and functions; the configuration-flag discipline, meaning naming conventions, guard patterns, and insertion points; the exact validation command and its expected output; the submission recipe, covering hardware type, resource entitlements, scheduling tags, package layer versions, and warm-start checkpoint configuration; the dead ends, meaning errors encountered in past iterations with their root causes and fixes, which serve as the system's scar tissue; and the proven strategies, meaning execution patterns that have consistently led to success, which serve as its muscle memory.

\subsection{Playbook Evolution as Text-Space Skill Optimization}
\label{sec:playbook-evolution}

The playbook evolves through a process analogous to text-space optimization~\citep{yang2026skillopt,lee2026metaharness}: every agent action during an iteration is logged as a step in the session trajectory, and when an iteration completes the trajectory is analyzed to extract knowledge that updates the playbook's natural-language content. Three kinds of knowledge are distilled. Dead ends capture error patterns together with their root causes and successful fixes, written as explicit avoidance instructions (for example, ``the runtime batch object exposes a feature under an internal field name that differs from its raw data-warehouse column name''); because each dead end pairs the error with its remedy, future iterations can self-heal. Pipeline recipes capture step-by-step procedures for the implementing, validating, training, and analyzing stages, distilled from successful trajectories, and specify which files to read, which commands to run, and which parameters to set. Submission configuration captures the infrastructure parameters learned through trial and error, such as hardware type, resource entitlements, package version, and scheduling tags, which are thereafter consumed automatically by every subsequent iteration.

One design insight, validated by examining 31 agent session transcripts, is that LLM agents consume procedural knowledge effectively as natural-language instructions. The playbook is a markdown document that the agent reads at session start; dead ends are written as ``DO NOT'' directives, and pipeline steps are written as numbered procedures. This natural-language interface is what allows for the measurable execution improvements documented in Section~\ref{sec:results}.

\subsection{Playbook Lifecycle and One-Shot Transfer Across Models}

\paragraph{One-shot bootstrap, then evolve.} The playbook evolves through a natural lifecycle that mirrors how a human researcher becomes proficient with a new model. In the first, interactive iteration, the human and the system work together: the human points to code, notebooks, and training configurations, the system summarizes them into an initial playbook, and error recovery still requires human guidance. Over the next several iterations, the system loads the playbook at session start and follows its proven patterns; dead ends accumulate, failure rates fall, and the human reviews at key checkpoints but intervenes less often. Once the playbook matures, its pipeline recipes are stable, its submission configuration is crystallized, and its dead-end catalog is comprehensive; autonomous mode becomes reliable, and the system executes the full loop without intervention, pausing only for the unavoidable asynchronous wait during training.

\begin{figure}[h]
    \centering
    \includegraphics[width=0.7\columnwidth]{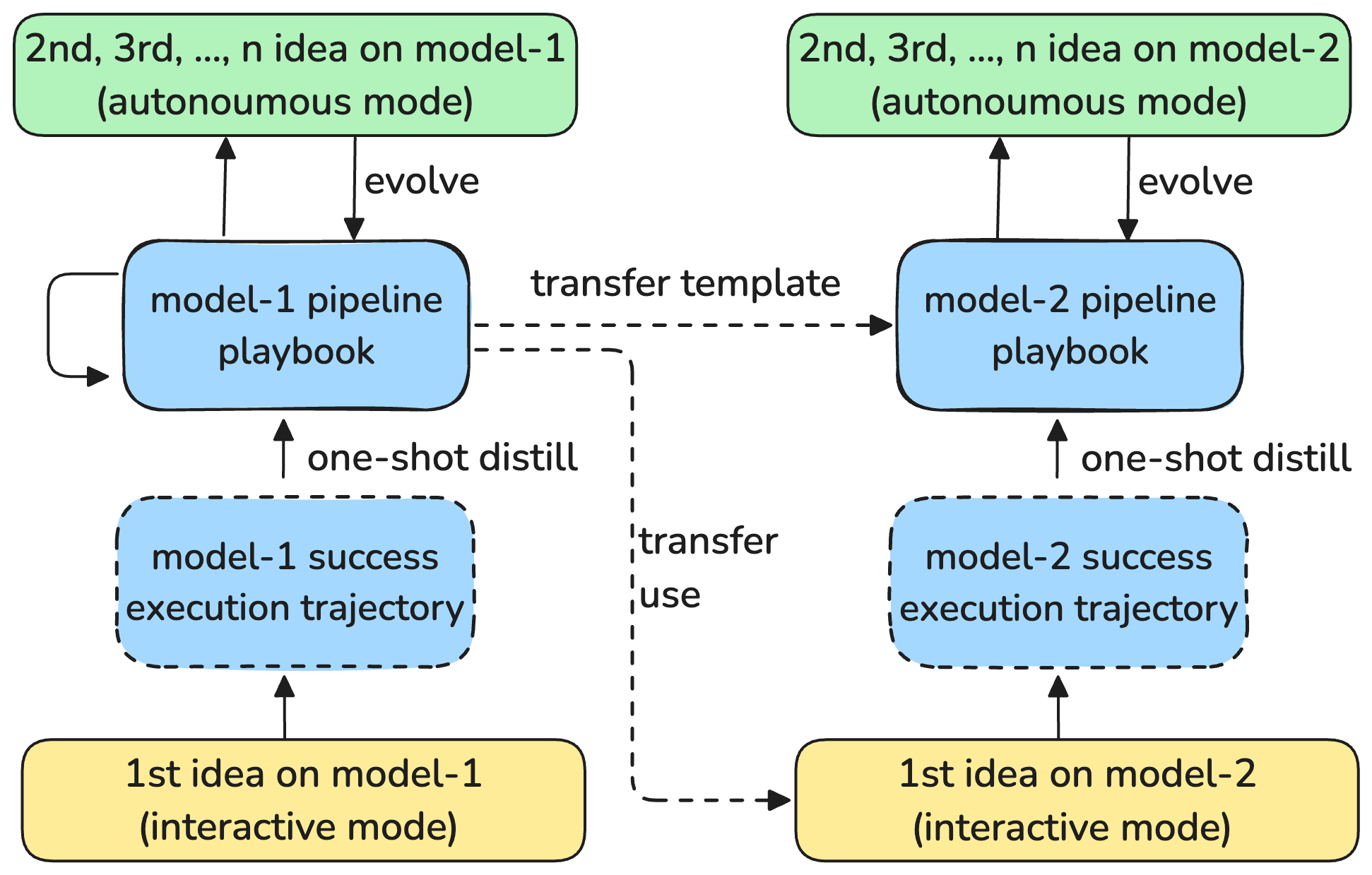}
    \caption{One-shot transfer playbook creation and evolution. The first model's playbook is distilled interactively over several iterations into a mature, well-structured template. Onboarding a later model is still interactive, but instead of rediscovering the structure the system only fills the template's slots with that model's specialized pointers. Every playbook, including transferred ones, then keeps evolving per iteration by accumulating new dead ends and crystallizing its submission configuration, which together let the system scale across many models.}
    \label{fig:oneshot-transfer}
\end{figure}

\paragraph{One-shot transfer across models.} A second innovation in this lifecycle is that the structure learned on the first model transfers to later models, so onboarding does not have to start from zero. The first time we run on a new model type, the interactive process above distills raw exploration into a well-structured playbook, and this structure, meaning the categories of operational knowledge and how they are organized, is model-agnostic even though its contents are not. Onboarding a later model is still interactive on its first run. What transfers is the structure: rather than rediscovering how a good playbook should be organized, the system reuses the first model's mature playbook as a template and, guided by the interactive session, fills its slots with the specialized pointers of the new model, namely its own key files, configuration conventions, validation command, submission recipe, and likely dead ends. We refer to this as one-shot transfer playbook creation, since a single mature template is enough to scaffold every subsequent model. It replaces a lengthy structural bootstrap with a focused interactive pass that populates a known scaffold, and it is the mechanism that lets Auto-RecSys scale across many models. Each transferred playbook still evolves independently thereafter, accumulating its own dead ends and crystallizing its own submission configuration as iterations proceed.

Together, this design directly targets the two demands of industry-scale research: it handles the operational complexity of each model by capturing hard-won, model-specific knowledge in a structured playbook, and it scales across many models by transferring that structure so every new model starts from a proven scaffold rather than from scratch.

\subsection{Self-Healing Through Dead End Avoidance}

Another strength of Auto-RecSys is its ability to self-heal. When the system encounters an error, it records not only the error but also the fix that resolved it, so that if a similar pattern appears in a later iteration the system applies the known fix automatically instead of escalating to the human. Once it learned that one GPU generation caused recurring hardware job failures, for instance, it defaulted to a more stable generation in all later submissions. Once it hit a package layer version mismatch, it recorded the compatible version and reused it. And once it found that a checkpoint had expired, it began looking up the latest valid checkpoint on its own. This behavior requires no manual rule authoring and emerges from the dead-end recording mechanism combined with the agent's ability to follow natural-language fix instructions. Because each dead end records why the error occurred and how it was fixed, rather than merely that it occurred, the system reasons about causes instead of symptoms.

\section{Idea Evolution Loop}
\label{sec:idea-evolution}

\subsection{Ideation Grounded in Model Architecture}

Auto-RecSys grounds its ideation in the actual model architecture. The centralized memory maintains a model context that captures the model architecture, task heads, feature inventory, and enabled modules, among others. This knowledge grounding is important at industry scale, where models have complex and non-obvious interactions between components. Maintaining this context as a persistent knowledge base carries a further practical benefit: the system does not have to rediscover the model from scratch at the start of every session. Because industry-scale models span thousands of lines of code and configuration, re-reading and re-deriving the architecture each time would consume a large number of tokens. By reusing the memorized context and refreshing only the parts that have changed, the system avoids this repeated exploration and spends its token budget on ideation rather than on relearning what it already knows.

Ideation draws candidate ideas from several complementary sources, as illustrated in \Cref{fig:system-overview}. Researchers can contribute proposals directly, for example through a design document that describes a change they want evaluated. The system can also mine external literature, adapting techniques from recent papers to the model at hand. And it can brainstorm on its own from the model knowledge base, analyzing architecture gaps and applying common improvement patterns, such as embedding strategies, gating mechanisms, and auxiliary losses, to the signals present in the data but not yet captured by the model, the missing interactions between components, and the underutilized modules. Whatever their origin, all candidate ideas pass through the same downstream pipeline. The system filters them against experiment history to see which kinds of ideas have worked, which have failed, and how to avoid repeats while building on partial successes, and it then ranks them by expected metric impact, implementation complexity, regression risk, and novelty relative to prior experiments.

Every idea is specified concretely, with a testable hypothesis, references of knowledge base, the target files it would modify, etc. This concreteness serves two purposes. It acts as a quality safeguard that makes each idea directly actionable once it is handed to the execution agent that drives the Execution Evolution Loop. It also supports human oversight: because experiments at industry scale are resource intensive, every idea is recorded in an organized, human-readable markdown file, so that a domain expert can review it and judge whether it makes sense before any training resources are committed.

\subsection{Parallel Idea Execution: The Distributed Portfolio}

\begin{figure}[h]
    \centering
    \includegraphics[width=0.7\columnwidth]{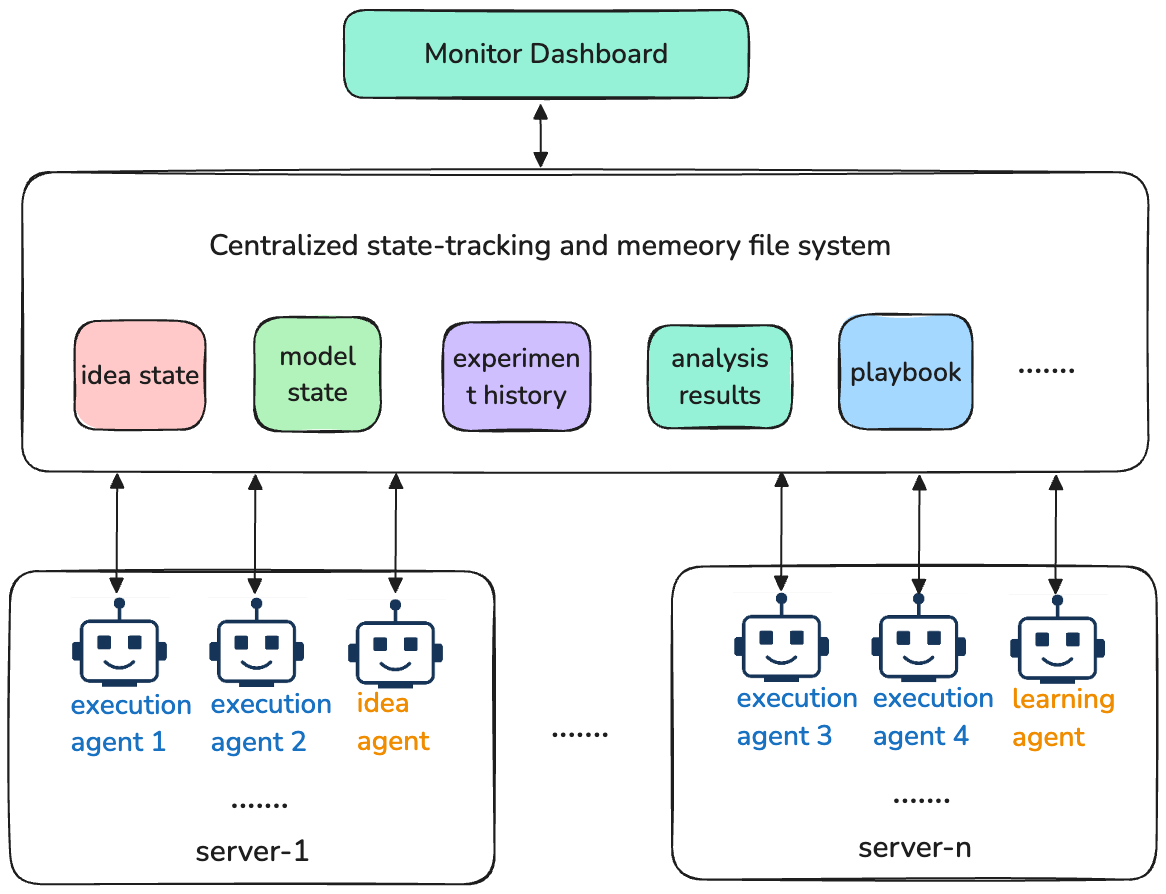}
    \caption{Parallel idea execution across servers. Multiple ideas run simultaneously at different lifecycle stages and synchronize through the centralized memory store. This distributed portfolio maintains research velocity despite multi-day training cycles while preserving shared learning and centralized oversight.}
    \label{fig:parallel-ideas}
\end{figure}

The long feedback loop of industry-scale training (hours to days per experiment) makes serial iteration impractical. Auto-RecSys addresses this by running multiple ideas in parallel as a \emph{distributed experiment portfolio}, as illustrated in \Cref{fig:parallel-ideas}.

Each idea maintains its own state file, so there is no contention between parallel ideas. All ideas on the same model share a common baseline with aligned training date ranges, ensuring fair A/B comparison. The global registry tracks which agent session is working on which idea, preventing conflicts and enabling the dashboard to display a unified view of all active experiments.

This portfolio approach transforms the throughput equation: while each individual experiment still takes days, the system as a whole processes multiple experiments per week. The researcher's role shifts from executing one idea at a time to \emph{managing a portfolio} of concurrent explorations.

\subsection{Learning from Outcomes}

When an idea's experiment completes, the analysis computes performance differences against the baseline and assigns a verdict of positive, neutral, or negative. The verdict, together with structured lessons learned, is appended to the model's experiment history, an append-only log that serves as the long-term memory of the Idea Evolution Loop. Later ideation sessions query this history to deduplicate against ideas already tried, to combine partial successes into stronger compound ideas (for example, pairing two features that each helped independently), to prune idea categories that have consistently failed on the model, and to target unexplored regions of the idea space.

\section{Persistence and Robustness}
\label{sec:persistence}

Robustness is not optional at industry scale; it is a hard requirement. A single agent session may consume a large number of tokens, and a single training job may consume hundreds of GPU-hours. Losing the context of a failed experiment, or repeating a configuration already known to be broken, wastes resources that could have gone toward productive exploration. Auto-RecSys addresses this through a carefully designed persistence architecture.

\subsection{Centralized Memory Store}

All system state resides on a centralized memory store accessible from each development server. This design choice provides several advantages for our use case: automatic cross-server synchronization, human-readable state files (JSON/JSONL), and natural backup through the storage provider's infrastructure.

The storage layout provides per-model, per-idea isolation:

\begin{lstlisting}[style=filesystem]
state/
  registry.json                 # Global: active ideas across servers
  <model>/
    active/<idea>.json          # In-flight idea state (mutable)
    completed/<idea>.json       # Archived (read-only)
    baselines.json              # Shared baselines with lifecycle
    experiment_history.jsonl    # Append-only log
    idea_backlog.json           # Ranked queue

playbooks/
  <model>.md                    # Human-readable recipe
  <model>_meta.json             # Iteration metadata

ideas/
  <model>/<idea>.md             # Per-idea details

knowledge_base/
  <model>/feature.md
  <model>/arch.md               # Model-specific knowledge

sessions/
  index.json                    # Session -> model/idea lookup
  <session_id>.jsonl            # Trajectory backup
\end{lstlisting}

To keep this shared state reliable, the system treats code as the harness for operationally sensitive state. Deterministic scripts read and write the registry, the per-idea state files, and the history through structured data formats with fixed schemas, rather than letting the LLM edit them freely. This pairing of code with a structured data format enforces precision on the state that must be exact, in contrast to the memory and knowledge layers, such as the playbooks and the model knowledge base, which are deliberately kept as human-readable natural language for the agent to reason over.

A further benefit of this structured representation is observability. Because all system state lives in well-defined, machine-readable files, it can be rendered directly into a centralized dashboard that gives the researcher a single, unified view of the whole system: every model, every in-flight idea and its current state, and the status of the training jobs running across all servers. This turns oversight of a large, distributed, multi-day experiment portfolio into a straightforward monitoring task and substantially improves the human-computer interaction of the system, which would be far harder to build if the same state were scattered across free-form logs or embedded in natural-language memory.

Two further mechanisms guard against data loss. The experiment history is stored as append-only JSONL, where each line is independently parseable, so corruption of one entry cannot affect the others. And every agent action is logged to a session trajectory file, so that a new session, possibly on a different server, can reconstruct what the previous session did; this trajectory log is the primary recovery mechanism after a session failure, as described next.

\subsection{Session Recovery and Multi-Server Handoff}

When a new agent session starts, the orchestrator runs a recovery protocol. It first reads the global registry to discover every active idea across all servers, then reads the state file of each active idea for the target model. If the current session identifier differs from the one on record, it reads the previous session's trajectory log to reconstruct context. For any idea in the \textsc{training} state it immediately polls the remote job for completion, and it finally presents a concise summary and routes to the appropriate next action.

This protocol enables seamless handoff between servers. A researcher can start an experiment on one server, submit training, close the session, and resume analysis on another server the next day. The system detects the server change, reads the trajectory from the memory store, polls the completed training job, and continues with analysis, all without the researcher needing to recall what happened.

\section{Evaluation and Observed Effectiveness}
\label{sec:results}

We have tested Auto-RecSys on several industry-scale recommendation models.

\subsection{Human Bandwidth per Idea}

A full comparison of end-to-end cycle time is not the right way to measure this system. The wall-clock duration of an experiment is dominated by training, which Auto-RecSys does not shorten, since an industry-scale run still takes days no matter how it is launched, and overall cycle time also depends on factors outside the system's control such as cluster queueing and baseline availability. The metric the system genuinely moves is the \emph{human bandwidth} each idea consumes, meaning the amount of hands-on researcher attention needed to carry an idea from proposal to analyzed result.

Under the manual process, a single idea of comparable complexity requires hours to days of active human involvement, spread across implementation, validation, submission, monitoring, failure recovery, and analysis. Auto-RecSys exposes a tradeoff between human control and human effort through its two operating modes. In human-in-the-loop mode, the researcher stays in control at two review checkpoints, idea selection and implementation review, and delegates the remaining execution, meaning validation, submission, monitoring, failure recovery, and analysis, to the system; for an idea of the same complexity, this reduces hands-on effort to minutes. In fully autonomous mode, the checkpoints are skipped and human involvement falls to a minimum, limited to selecting ideas and responding when the system escalates an issue it cannot resolve on its own; this minimizes human time but risks pursuing weaker ideas and spending training resources on them, so we favor it only once the playbook is mature and the candidate ideas are low-risk. Either way, the benefit is not faster training but substantially higher researcher throughput per unit of attention: on the same hands-on allotment, the attention that once covered a single idea now covers more than a dozen, and running several ideas in parallel is what makes the distributed portfolio practical.

\subsection{Playbook Evolution and Execution Reliability}

We analyze execution reliability across 31 unique experiment iterations on one model. We choose this model because it is a representative model and, importantly, because it underwent a baseline shift during the evaluation period. This shift lets us observe the system's learning and evolving behavior in two regimes: as the playbook first matures on a stable baseline, and as it recovers after changes to the underlying model. We introduce the following metrics derived from the session trajectory logs:

\begin{itemize}
    \item \textbf{Session log step.} Each autonomous iteration records a sequence of timestamped actions in the trajectory log, including implementation decisions, observations, job submissions, error recoveries, and analysis results. Each entry constitutes one \emph{session log step}, representing a discrete agent action within the iteration lifecycle.

    \item \textbf{Major fix step.} A session log step spent recovering from an \emph{operational} failure, such as a failed or resubmitted training job, a wrong hardware or entitlement setting, a package layer version mismatch, a metadata correction, or a baseline refresh. We deliberately exclude the reasoning and debugging that occur while implementing an idea in code, because that effort is tied to the complexity of the idea and the capability of the underlying model rather than to playbook maturity, and we count it as productive work instead. Restricting the metric to operational failures lets it reflect what the playbook actually controls, namely the failures that a more knowledgeable playbook could have avoided.

    \item \textbf{Zero-fix rate.} The fraction of iterations that complete with zero major fix steps, that is, iterations that run from idea to analysis without any operational error recovery. This measures end-to-end operational reliability.

    \item \textbf{Error category.} A class of operationally related failures sharing a common root cause (e.g., ``GPU hardware instability'' or ``build date parameter conventions''). The playbook records individual errors as dead ends; we group them into categories to analyze whether entire classes of failures are eliminated across phases.
\end{itemize}

\begin{figure*}[t]
    \centering
    \begin{subfigure}[t]{0.58\textwidth}
        \centering
        \includegraphics[width=\textwidth]{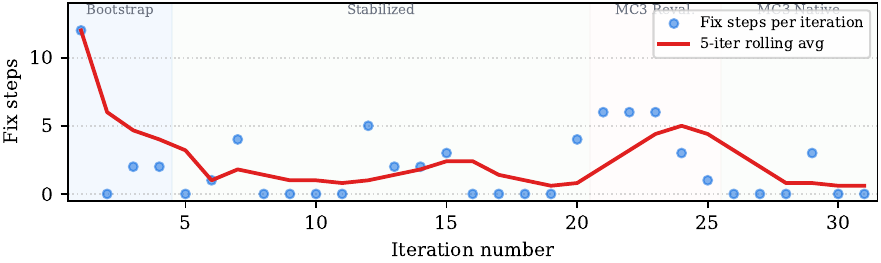}
        \caption{Per-iteration major (operational) fix steps with a 5-iteration rolling average across 31 unique iterations. The rolling average drops during stabilization (iterations 5 to 20), spikes sharply at the baseline transition (iterations 21 to 25), and recovers to its best levels to date in the post-transition phase (iterations 26 to 31).}
        \label{fig:fix-timeline}
    \end{subfigure}
    \hfill
    \begin{subfigure}[t]{0.38\textwidth}
        \centering
        \includegraphics[width=\textwidth]{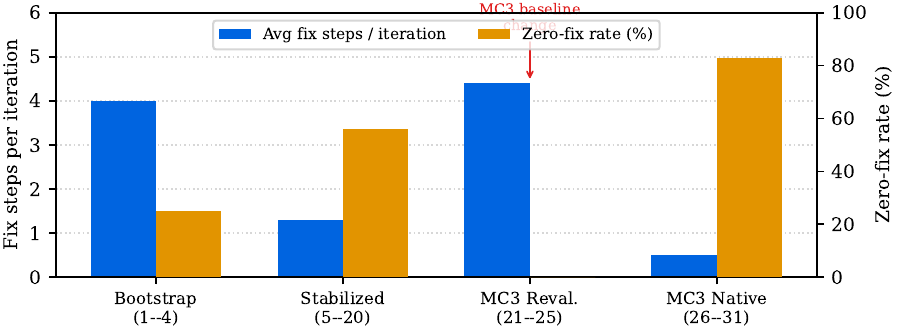}
        \caption{Phase-level summary. Major fixes per iteration (blue, left axis) and zero-fix rate (orange, right axis). The baseline transition resets both metrics to bootstrap levels before the playbook recovers.}
        \label{fig:phase-bars}
    \end{subfigure}
    \caption{Execution reliability evolution across 31 unique iterations on the same model. Data sourced from session trajectory logs and playbook metadata.}
    \label{fig:playbook-evolution}
\end{figure*}

\paragraph{Evolution trajectory.} Figure~\ref{fig:playbook-evolution} visualizes the per-iteration major fix step counts and phase-level statistics. There is a baseline transition at iteration 21, when the baseline changed from a standard architecture to a combined configuration that integrated graph-mode compilation, a new embedding module, and task adapters; this altered the submission infrastructure, requiring different hardware, entitlements, package layer versions, and upstream revisions, and created a natural experiment in playbook adaptability. The rolling average shows a clear learn, regress, and recover cycle centered on this transition. During stabilization (iterations 5 to 20), the playbook absorbed hardware choices, submission parameters, build conventions, and input-naming rules, reducing major fixes from 4.0 to 1.3 per iteration. When the new baseline was introduced at iteration 21, five consecutive iterations required operational recovery due to wrong hardware and entitlements for the new configuration, adapter injection failures, package layer version mismatches, and upstream revision conflicts. After absorbing these transition-specific lessons, the post-transition phase (iterations 26 to 31) reached 0.5 major fixes per iteration, with 5 of its 6 iterations requiring no operational fix at all, surpassing even the pre-transition stabilized performance.

\paragraph{Error category analysis.} A breakdown of failures by category reveals why the playbook improves: errors are categorical rather than random. Each category, such as GPU hardware instability, missing resource tags, incorrect build date parameters, input-naming conventions, and package layer version mismatches, appears in one or two phases, is recorded as a dead end, and then disappears from the phases where the playbook has absorbed the fix. The transition revalidation phase introduces two new categories, package layer mismatches and publish infrastructure failures, both of which are absent from the post-transition phase. The only post-transition error is a genuinely novel code bug, a type-inference failure in graph compilation, that no prior experience could have prevented.

\paragraph{How the playbook influences agent behavior.} Examination of all 31 agent session transcripts reveals that the playbook's influence operates primarily through natural language. The agent reads the playbook markdown file at session start and follows its instructions: the dead-ends table (19 entries with explicit ``DO NOT'' directives) prevents known errors, the proven-strategies table guides implementation patterns, and the pipeline recipes specify exact tool sequences. The agent seldom queries numerical metadata or scoring data; it prefers to consume operational knowledge as natural-language instructions.

This validates the choice of natural-language knowledge distillation as the interface for an LLM agent's procedural memory. A dead-end description such as ``the runtime batch object exposes a feature under an internal field name that differs from its raw data-warehouse column name'' is directly actionable by the agent, whereas a numerical confidence score would require an additional interpretation step. The playbook's effectiveness comes from writing errors and fixes in the same language the agent reasons in.

The playbook crystallized a per-stage execution pipeline with specific tool sequences: read calls on 6 key files during implementation, the toy-train command for validation, build-based submission with 9 pinned parameters for training, and a metrics-fetching call followed by a baseline-comparison call for analysis. This pipeline evolved from the 49 dead ends and 17 error-fix patterns accumulated in the playbook.

\paragraph{Concrete self-evolution examples.} Examination of the raw agent session transcripts (46 sessions across more than ten development servers) reveals five distinct self-evolution mechanisms:

\begin{itemize}
    \item \textbf{Dead-end avoidance.} During bootstrapping, the system learned through three failed jobs that one GPU generation caused recurring hardware job failures, recorded this as a dead end, and thereafter selected a stable generation automatically; a later iteration explicitly loaded the lesson that flash attention was broken on that hardware and a manual fallback should be used.
    \item \textbf{Pipeline config crystallization.} The playbook's submission configuration accumulated hard-won parameters, including hardware type, package layer version, resource entitlements, scheduling tags, resource-partition flags, ownership metadata, the build target, the model type, etc, each learned from a specific failure and then consumed by every subsequent iteration.
    \item \textbf{Validation-skip pattern.} When a development server lacked a GPU, the playbook guided the system to skip local toy-train validation and go straight to remote training submission, avoiding a failed validation attempt and unnecessary human intervention each time.
    \item \textbf{Self-diagnosed infrastructure improvement.} During multi-day monitoring, the system found that background monitor agents were dying silently after roughly three to five hours because each polling cycle appended tool results until the LLM context window overflowed. It autonomously designed a cron-based replacement in which each tick is a fresh prompt with no context accumulation, implemented it, and submitted a code change that upgraded the orchestrator, improving its own infrastructure rather than just the model.
    \item \textbf{Emergent meta-pattern recognition.} After finalizing the sixth consecutive pre-transition positive that failed on the new baseline, the agent synthesized, without being prompted, that the new baseline already captured the signals these features provided, making them redundant and even harmful; this observation was recorded as a learning and directly drove the pivot toward experiments native to the new baseline.
\end{itemize}

Across these mechanisms, the common effect is that the playbook's self-evolution steadily eliminates redundant exploration and error recovery. The per-iteration compute cost of the agent is already modest relative to GPU training time, minutes of reasoning against days of training, but trimming this wasted effort still lowers the agent's own token cost and, more importantly, frees researcher attention for new ideas.

\subsection{Robustness in Practice}

The persistent state machine proved beneficial for industry-scale operation. Examination of the session transcripts reveals three concrete recovery patterns:

\paragraph{Cross-server context recovery.} Consider a representative case. The agent working an idea got stuck, finished implementing the change, and then crashed almost immediately when its development server's lease expired, before it could validate or submit training. Another agent on a different server later resumes the same idea. Because each server checks out independently, the code from the previous session is not present in the new checkout. The resuming agent detects this from the local commit history (``the commit from the previous session doesn't exist in this checkout; it was working on a different server'') and reads the trajectory log to reconstruct the context of what was done. It then fetches the draft diff from structured state file (via \textsc{get\_diff\_num(idea)}) that the earlier session had submitted to the shared code-review system and applies it directly, then validates, submits training, and finalizes the iteration, all without human guidance. Because every code change is published as a draft diff rather than left only in a local checkout, any agent on any server can resume an in-flight idea directly from the shared state and continue working from there, which makes cross-server handoff seamless and avoids the token cost of regenerating work.

\paragraph{Autonomous execution depth.} In the most autonomous session observed, the agent executed 970 consecutive log entries (110 tool calls) with zero human intervention, diagnosing failed training jobs, identifying a root cause in a fused-kernel import within a shared operator library, rebuilding package layers, and resubmitting experiments. These ratios show that the system operates autonomously for extended periods, with human input limited to high-level direction.

\paragraph{Adaptive workflow modification.} After one feature-gating experiment failed four times at the publish step due to an ahead-of-time compilation failure in a shared ranking module, the system adapted its workflow, switching to a training flow that skipped the failing publish step that had killed the previous runs. The next attempt ran clean on the first try. This shows the system modifying not just parameters but the execution workflow itself in response to accumulated failure experience.

\section{Related Work}

\label{sec:related}

\paragraph{Autonomous research systems.}

Recent systems increasingly automate substantial portions of the scientific workflow, including hypothesis generation, implementation, experimentation, analysis, and paper writing. The AI Scientist~\citep{aiscientist} studies end-to-end automated scientific discovery, while AutoResearch~\citep{autoresearch} demonstrates rapid serial experiments on small-scale problems. FARS~\citep{fars} produces research artifacts at scale on self-contained academic tasks, whereas EvoScientist~\citep{evoscientist2026} and Dr.~Claw automate broader workflows spanning literature review, ideation, experimentation, analysis, and paper generation. These systems establish the potential of autonomous research when experiments are relatively bounded and feedback is available within short iteration cycles. Auto-RecSys studies a different operating regime: industry-scale recommendation models whose training takes hours or days, depends on complex infrastructure, and requires comparison against continuously evolving baselines. These constraints shift autonomous research from rapid serial iteration toward distributed parallel exploration, persistent state management, and recoverable execution across sessions and servers.

\paragraph{Agent harnesses and experiment orchestration.}

An agent harness is the software and context layer surrounding a language model, including its tools, APIs, memory, validators, execution loops, and feedback channels~\citep{lee2026metaharness,ning2026codeharness}. ReAct~\citep{yao2023react} established the interleaving of reasoning and environment interaction, while SWE-agent~\citep{yang2024sweagent} demonstrated that the agent--computer interface can substantially affect task performance. DSPy~\citep{khattab2023dspy} introduced the programmatic composition and optimization of language-model pipelines. Ning et al.~\citep{ning2026codeharness} formalize code as an agent harness with three defining properties: executability, verifiability, and statefulness, and identify automated harness evolution as an open research direction. Meta-Harness~\citep{lee2026metaharness} optimizes harness configurations using execution outcomes and traces, while ADAS~\citep{hu2024adas} searches over agentic system designs through a meta-agent that programs new architectures. Auto-RecSys extends these ideas from inference-time task execution to the full lifecycle of long-running model experimentation. Its harness schedules parallel experiments, preserves state across sessions and failures, coordinates infrastructure interactions, and accumulates reusable execution knowledge. Unlike conventional experiment-management platforms that primarily track runs, artifacts, and metrics, Auto-RecSys autonomously executes experiments, recovers from failures, and uses prior outcomes to guide subsequent actions.

\paragraph{Procedural memory, skills, and self-improvement.}

Agent memory and skills enable language-model agents to improve through accumulated experience rather than treating each task independently. Reflexion~\citep{shinn2023reflexion} stores verbal feedback for episodic improvement, Voyager~\citep{wang2023voyager} maintains a growing library of executable skills, and MemGPT~\citep{packer2023memgpt} organizes information through tiered memory. Agent Workflow Memory (AWM)~\citep{wang2024agent} induces reusable workflows from prior trajectories and selectively retrieves them to guide future actions, closely relating to the pipeline recipes maintained by Auto-RecSys. Evo-Memory~\citep{wei2025evo} studies test-time memory evolution over sequential task streams, where agents continually retrieve, integrate, and update experience across interactions. Recent work further formalizes skills as reusable procedural artifacts containing tool policies, applicability conditions, routines, and failure modes~\citep{li2026skillsbench,jiang2026sokskills}. Trace2Skill~\citep{ni2026trace2skill} distills transferable skills from trajectories, EvoSkill~\citep{alzubi2026evoskill} discovers skills through failure-guided evolutionary search, and SkillOpt~\citep{yang2026skillopt} formulates skill learning as constrained text-space optimization with validation and rejected-edit mechanisms. GEPA~\citep{agrawal2025gepa} similarly demonstrates improvement through reflective evolution of textual instructions. Auto-RecSys combines these strands at the system level: execution trajectories update model-specific playbooks that improve \emph{how} experiments are conducted, while experimental outcomes update research history that improves \emph{what} is attempted next. Unlike prior methods studied primarily within individual task episodes or benchmark task streams, these two forms of memory are coupled to persistent, multi-server orchestration and multi-day industry-scale experiments.

\section{Discussion and Future Work}
\label{sec:discussion}

\paragraph{Scaling the idea loop with proxy models.} A remaining bottleneck is the long training time per experiment. Future work will explore proxy models (smaller-scale replicas of industry-scale models) for rapid idea screening. If small-scale results transfer reliably to full-scale, the system could screen dozens of ideas quickly, then focus on full-scale training only in the most promising candidates. This would bring the Idea Evolution Loop closer to AutoResearch's rapid serial iteration while maintaining the rigor of industry-scale evaluation for the final candidates.

\paragraph{Cross-model knowledge transfer.} Currently each model maintains independent playbooks and experiment histories. A natural extension is cross-model learning: insights like ``gating mechanisms help multi-task models'' or ``embedding dimension sweeps have diminishing returns above 128'' could transfer across models with similar architectures. This would accelerate the Idea Evolution Loop for new models and reduce the cold-start problem for playbooks.

\paragraph{Validation-gated playbook updates.} A notable gap relative to SkillOpt~\citep{yang2026skillopt} is the absence of a formal validation gate for playbook updates. SkillOpt's ablation shows that removing the validation gate degrades performance. Auto-RecSys currently accepts playbook updates unconditionally based on agent judgment. This has not caused observable regressions because dead ends record actual failures (they are correct by construction), but a lightweight validation mechanism (e.g., verifying that a new dead-end entry does not conflict with existing proven strategies) could further improve robustness, particularly as playbooks grow in size and complexity.

\paragraph{Adaptive human-in-the-loop.} The current interactive/autonomous toggle is binary. A more nuanced approach would have the system request human input selectively, based on a calibrated confidence estimate: high-confidence, low-computational cost decisions proceed autonomously, while new or high-impact decisions (e.g., first experiment on a new model, idea with high regression risk) trigger human review.

\paragraph{Scaling to more researchers and models.} The current system is designed for a single researcher managing a portfolio of models. Extending to team-scale operation introduces new challenges: shared idea backlogs, collaborative experiment history, and multi-user conflict resolution. The per-model isolation architecture provides a natural starting point for this extension.

\section{Conclusion}
\label{sec:conclusion}
We presented Auto-RecSys, an autonomous research harness designed to accelerate recommendation model innovation in industry-scale environments. Auto-RecSys addresses the long feedback loops and operational complexity of large-scale experimentation through a centralized memory substrate that supports distributed execution, persistent state management, cross-server recovery, and cumulative knowledge reuse. Building on this substrate, its dual evolution architecture continuously improves both execution and research: the Execution Evolution Loop distills operational experience into model-specific natural-language playbooks, while the Idea Evolution Loop coordinates parallel exploration and uses experimental outcomes to guide subsequent hypotheses. In evaluation, Auto-RecSys substantially reduces human effort per experiment cycle and decreases major execution fixes as its playbooks mature, from 4.0 to 0.5 fixes per iteration across 31 unique iterations, while rapidly adapting to changes in the underlying baseline architecture. These results demonstrate that autonomous research can be extended beyond small, rapidly evaluated tasks to complex industry-scale systems by equipping agents with a robust, persistent, and continually evolving execution harness.

\section{Acknowledgments}
\label{sec:acknowledgments}
We would like to thank Shilin Ding for actionable feedback and strategic guidance that strengthened the design and positioning of this work.

\bibliographystyle{assets/plainnat}
\bibliography{paper}

\end{document}